\documentclass[11pt]{article}
\usepackage[utf8]{inputenc}
\usepackage{tikz}
\usepackage{times}
\usepackage{url}
\usepackage{latexsym}
\usepackage{amssymb}
\usepackage{amsmath}
\usepackage{multibib}
\usepackage{graphicx}
\usepackage{float}
\usepackage{graphics}
\usepackage{coling2016}
\usepackage{booktabs}
\usepackage[]{algorithm2e}
\usepackage{multirow}
\usepackage{tabularx} 
\usepackage{changepage}
\usepackage{enumitem}
\usepackage[english]{babel}
\usetikzlibrary{bayesnet}

\title{`Who would have thought of that!': A Hierarchical Topic Model for Extraction of Sarcasm-prevalent Topics and Sarcasm Detection}
\author{\begin{tabular}{ccc}
Aditya Joshi$^{1,2,3}$ & &Prayas Jain$^{4}$\\
Pushpak Bhattacharyya$^{1}$&&Mark James Carman$^{2}$\\
\end{tabular}\\
\begin{tabular}{ccc}
 \multicolumn{3}{c}{$^{1}$Indian Institute of Technology Bombay, India, $^{2}$Monash University, Australia}\\
\multicolumn{3}{c}{$^{3}$IITB-Monash Research Academy, India, $^{4}$IIT-BHU (Varanasi), India, }\\
\multicolumn{3}{c} {\tt \{adityaj, pb\}@cse.iitb.ac.in}, \tt prayas.jain.cse14@iitbhu.ac.in\\
\multicolumn{3}{c}{\tt mark.carman@monash.edu}\\
\end{tabular}
}

\begin{document}

\maketitle

\begin{abstract}
Topic Models have been reported to be beneficial for aspect-based sentiment analysis. This paper reports a simple topic model for sarcasm detection, a first, to the best of our knowledge. Designed on the basis of the intuition that sarcastic tweets are likely to have a mixture of words of both sentiments as against tweets with literal sentiment (either positive or negative), our hierarchical topic model discovers sarcasm-prevalent topics and topic-level sentiment. Using a dataset of tweets labeled using hashtags, the model estimates topic-level, and sentiment-level distributions. Our evaluation shows that topics such as `work', `gun laws', `weather' are sarcasm-prevalent topics. Our model is also able to discover the mixture of sentiment-bearing words that exist in a text of a given sentiment-related label. Finally, we apply our model to predict sarcasm in tweets. We outperform two prior work based on statistical classifiers with specific features, by around 25\%. 
\end{abstract}
\section{Introduction}
\textit{This paper will be presented at the 3rd ExPROM workshop at COLING 2016. \url{http://www.cse.unt.edu/exprom2016/}}\\
Sarcasm detection is the computational task of predicting sarcasm in text. Past approaches in sarcasm detection rely on designing classifiers with specific features (to capture sentiment changes or incorporate context about the author, environment, etc.) \cite{joshiacl,wallace-EtAl:2014:P14-2,Rajadesingan:2015:SDT:2684822.2685316,bamman2015contextualized}, or model conversations using the sequence labeling-based approach by ~\newcite{joshiconll}. Approaches, in addition to this statistical classifier-based paradigm are: deep learning-based approaches as in the case of~\newcite{silvioconll} or rule-based approaches such as ~\newcite{riloffemnlp,joshiwassa}. 

This work \textit{employs a machine learning technique that, to the best of our knowledge, has not been used for computational sarcasm. Specifically, we introduce a topic model for extraction of sarcasm-prevalent topics and as a result, for sarcasm detection}. Our model based on a supervised version of the Latent Dirichlet Allocation (LDA) model~\cite{lda} is able to discover clusters of words that correspond to sarcastic topics. The goal of this work is to discover sarcasm-prevalent topics based on sentiment distribution within text, and use these topics to improve sarcasm detection. The key idea of the model is that (a) some topics are more likely to be sarcastic than others, and (b) sarcastic tweets are likely to have a different distribution of positive-negative words as compared to literal positive or negative tweets. Hence, distribution of sentiment in a tweet is the central component of our model.

Our sarcasm topic model is learned on tweets that are labeled with three sentiment labels: literal positive, literal negative and sarcastic. In order to extract sarcasm-prevalent topics, the model uses three latent variables: a topic variable to indicate words that are prevalent in sarcastic discussions, a sentiment variable for sentiment-bearing words related to a topic, and a switch variable that switches between the two kinds of words (topic and sentiment-bearing words). Using a dataset of 166,955 tweets, our model is able to discover words corresponding to topics that are found in our corpus of positive, negative and sarcastic tweets.

We evaluate our model in two steps: a \textbf{qualitative evaluation} that ascertains sarcasm-prevalent topics based on the ones extracted, and a \textbf{quantitative evaluation} that  evaluates sub-components of the model. We also demonstrate how it can be used for sarcasm detection. To do so, we compare our model with two prior work, and observe a significant improvement of around 25\% in the F-score. 

The rest of the paper is organized as follows. Section ~\ref{sec:relwork} discusses the related work. Section ~\ref{sec:motiv} presents our motivation for using topic models for automatic sarcasm detection. Section~\ref{sec:model} describes the design rationale and structure of our model. Section~\ref{sec:expsetup} describes the dataset and the experiment setup. Section~\ref{sec:res} discusses the results in three steps: qualitative results, quantitative results and application of our topic model to sarcasm detection. Section~\ref{sec:concl} concludes the paper and points to future work.
\section{Related Work}
\label{sec:relwork}
Topic models are popular for sentiment aspect extraction. \newcite{asum} present an aspect-sentiment unification model that learns different aspects of a product, and the words that are used to express sentiment towards the aspects. In terms of using two latent variables: one for aspect and one for sentiment, they are related to our model. \newcite{arjun} use a semi-supervised model in order to extract aspect-level sentiment. The role of the supervised sentiment label in our model is similar to their work. Finally, \newcite{ratingdimensions} attempt to generate rating dimensions of products using topic models. However, the topic models that have been reported in past work have been for sentiment analysis in general. They do not have any special consideration to either sarcasm as a label or sarcastic tweets as a special case of tweets. The hierarchy-based structure (specifically, the chain of distributions for sentiment label) in our model is based on ~\newcite{joshi2016political} who extract politically relevant topics from a dataset of political tweets. The chain in their case is sentiment distribution of an individual and a group. 

Sarcasm detection approaches have also been reported in the past~\cite{teraarxiv,liebrecht2013perfect,29,joshiacl}.~\newcite{29} present a contextual model for sarcasm detection that collectively models a set of tweets, using a sequence labeling algorithm - however, the goal is to detect sarcasm in the last tweet in the sequence. The idea of distribution of sentiment that we use in our model is based on the idea of context incongruity. In order to evaluate the benefit of our model to sarcasm detection, we compare two sarcasm detection approaches based on our model with two prior work, namely by \newcite{buschmeier2014impact} and \newcite{liebrecht2013perfect}. \newcite{buschmeier2014impact} train their classifiers using features such as unigrams, laughter expressions, hyperbolic expressions, etc. \newcite{liebrecht2013perfect} experiment with unigrams, bigrams and trigrams as features. To the best of our knowledge, past approaches for sarcasm detection do not use topic modeling, which we do.
\section{Motivation}
\label{sec:motiv}
Topic models enable discovery of thematic structures in a large-sized corpus. The motivation behind using topic models for sarcasm detection arises from two reasons: (a) presence of sarcasm-prevalent topics, and (b) differences in sentiment distribution in sarcastic and non-sarcastic text. In context of sentiment analysis, topic models have been used for aspect-based sentiment analysis in order to discover topic and sentiment words~\cite{asum}. The general idea is that for a restaurant review, the word `spicy' is more likely to describe food as against ambiance. On similar lines, the discovery that a set of words belong to a sarcasm-prevalent topic - a topic regarding which sarcastic remarks are common - can be useful as additional information to a sarcasm detection system. The key idea of our approach is that some topics are more likely to evoke sarcasm than some others. For example, a tweet about working late night at office/ doing school homework till late night is much more probable to be sarcastic than a tweet on Mother's Day. A sarcasm detection system can benefit from incorporating this information about sarcasm-prevalent topics. The second reason is the difference in sentiment distributions. A positive tweet is likely to contain only positive words, a negative tweet is likely to contain only negative words. On the other hand, a sarcastic tweet may contain a mix of the two kind of words (for example, `\textit{I love being ignored}' where `\textit{love}' is a positive word and `\textit{ignored}' is a negative word), except in the case of hyperbolic sarcasm (for example `\textit{This is the best movie ever!}' where `\textit{best}' is a positive word and there is no negative word). Hence, in addition to sarcasm-prevalent topics, sentiment distributions for tweets also form a critical component of our topic model.
\begin{table*}[]
\centering
\begin{tabular}{ll}
\toprule
\multicolumn{2}{l}{\textbf{Observed Variables and Distributions}} \\
\midrule
$w$ & Word in a tweet \\
$l$ & Label of a tweet; takes values: positive, negative, sarcastic) \\
\multicolumn{2}{l}{\textbf{Distributions}}\\
$\eta_w$ & Distribution over switch values given a word w\\
\midrule
\multicolumn{2}{l}{\textbf{Latent Variables and Distributions}} \\
\midrule
$z$ & Topic of a tweet \\
$s$ & Sentiment of a word in a tweet; takes values: positive, negative \\
$is$ & Switch variable indicating whether a word is a topic word or a sentiment word; takes values: 0, 1   \\
\multicolumn{2}{l}{\textbf{Distributions}}\\
$\theta_l$ & Distribution over topics given a label l, with prior $\alpha$\\
$\phi_z$ & Distribution over words given a topic z and switch =0 (topic word), with prior $\gamma$\\
$\chi_s$ & Distribution over words given sentiment s and switch=1 (sentiment word), with prior $\delta_1$\\
$\chi_{sz}$ & Distribution over words given a sentiment s and topic z and switch=1 (sentiment word), with prior $\delta_2$\\
$\psi_l$ & Distribution over sentiment given a label l and switch =1 (sentiment word), with prior $\beta_1$\\
$\psi_{zl}$ & Distribution over sentiment given a label l and topic z and switch =1 (sentiment word), with prior $\beta_2$\\
\bottomrule
\end{tabular}
\caption{Glossary of Variables/Distributions used}
\label{tab:vars}
\end{table*}

\section{Sarcasm Topic Model}
\label{sec:model}

\subsection{Design Rationale}
Our topic model requires sentiment labels of tweets, as used in ~\newcite{Ramage:2009:LLS:1699510.1699543}. This sentiment can be positive or negative. However, in order to incorporate sarcasm, we re-organize the two sentiment values into \underline{three}: literal positive, literal negative and sarcastic. The observed variable $l$ in our model indicates this sentiment label. \textit{For sake of simplicity, we refer to the three values of $l$ as positive, negative and sarcastic, in rest of the paper.} 

Every word $w$ in a tweet is either a topic word or a sentiment word. A topic word arises due to a topic, whereas a sentiment word arises due to combination of topic and sentiment. This notion is common to several sentiment-based topic models from past work~\cite{asum}. To determine which of the two (topic or sentiment word) a given word is, our model uses three latent variables: a tweet-level topic label $z$, a word-level sentiment label $s$, and a switch variable $is$. Each tweet is assumed to have a single topic indicated by $z$. The single-topic assumption is reasonable considering the length of a tweet. At the word level, we introduce two variables $is$ and $s$. For each word in the dictionary, $is$ denotes the probability of the word being a  topic word or a sentiment word. Thus, the model estimates three sets of distributions: (A) Probability of a word belonging to topic ($\phi_z$) or sentiment-topic combination ($\chi_{sz}$), (B) Sentiment distributions over label and topic ($\psi_{zl}$), and (C) Topic distributions over label ($\theta_l$). The switch variable $is$ is sampled from $\eta_w$, the probability of the word being a topic word or a sentiment word. We thus allow a word to be either a topic word or a sentiment word.\footnote{Note that $\eta_w$ is not estimated during the sampling but learned from a large-scale corpus, as will be described later.}

\subsection{Plate Diagram}

\tikzstyle{init} = [pin edge={to-,thin,black}]
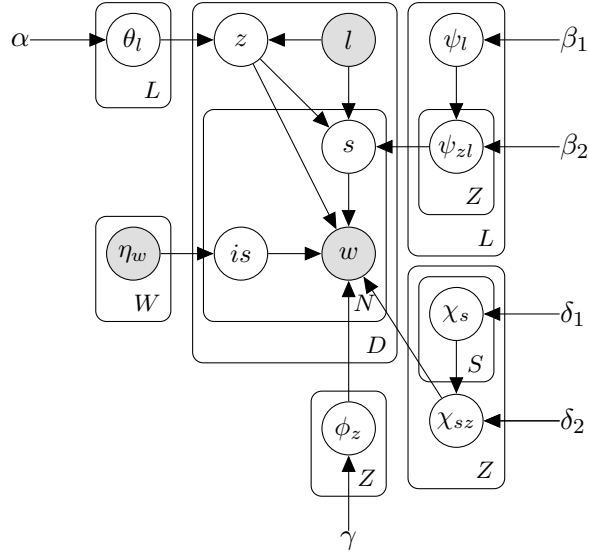
\begin{figure}[!ht]
\centering
\begin{tikzpicture}

  \node[obs] (l) {$l$} ;
  \node[latent, left=of l] (z) {$z$} ;
  \node [latent, below=of l] (s) {$s$} ;
  \node [obs, below=of s] (w) {$w$} ;
  \node [latent, left=of w] (is) {$is$} ;
  \node [obs, left=of is] (eta_w) {$\eta_w$} ;
  \node [latent, left=of z] (theta_l) {$\theta_l$} ;
  \node [latent, below=of w, yshift=-0.9cm] (phi_z) {$\phi_z$} ;
  \node[const, left=of theta_l](alpha){$\alpha$};
  \node[latent, right=of l] (psi_l) {$\psi_l$} ;
  \node[latent, below=of psi_l] (psi_zl) {$\psi_{zl}$} ;
  \node[latent,right=of w, yshift=-0.8cm] (chi_s) {$\chi_s$} ;
  \node[latent, below=of chi_s] (chi_sz) {$\chi_{sz}$} ;
  \node[const, right=of psi_l](beta_1){$\beta_1$};
  \node[const, right=of psi_zl](beta_2){$\beta_2$};
  \node[const, right=of chi_s](delta_1){$\delta_1$};
  \node[const, right=of chi_sz](delta_2){$\delta_2$};
   \node[const, below=of phi_z](gamma){$\gamma$};

  \edge {l} {z,s} ;
  \edge {z} {s,w} ;
  \edge {s} {w} ;
  \edge {is} {w} ;
  \edge {eta_w} {is} ;
  \edge {theta_l} {z} ;
  \edge {phi_z} {w} ;
  \edge{alpha}{theta_l};
  \edge {psi_l} {psi_zl} ;
  \edge {psi_zl} {s} ;
  \edge {chi_s} {chi_sz} ;
  \edge{chi_sz} {w} ;
  
  \edge {gamma}{phi_z} ;
  \edge {delta_2} {chi_sz} ;
  \edge {delta_1} {chi_s} ;
  \edge {delta_2} {chi_sz} ;
  \edge {beta_2} {psi_zl} ;
  \edge {beta_1} {psi_l} ;

  \plate {n_plate} {(s)(is)(w)} {$N$} ;
  \plate {d_plate} {(z)(l)(n_plate)} {$D$} ;
  \plate {eta_plate} {(eta_w)} {$W$} ;
  \plate {theta_plate} {(theta_l)} {$L$} ;
  \plate {phi_plate} {(phi_z)} {$Z$} ;
  \plate {psi_zl_plate} {(psi_zl)} {$Z$};
  \plate {psi_l_plate} {(psi_l)(psi_zl_plate)} {$L$};
 \plate {chi_s_plate} {(chi_s)} {$S$} ;
 \plate {chi_sz_plate} {(chi_sz)(chi_s_plate)} {$Z$} ;
\end{tikzpicture}
\caption{Plate Diagram of Sarcasm Topic Model}
\label{fig:diag}
\end{figure}
Our sarcasm topic model to extract sarcasm-prevalent topics is based on supervised LDA~\cite{lda}. Figure ~\ref{fig:diag} shows the plate diagram while Table~\ref{tab:vars} details the variables and distributions in the model. Every tweet consists of a set of observed words $w$ and one tweet-level, observed sentiment label $l$. The label takes \textbf{three} values: positive, negative or sarcastic. The third label value `sarcastic' indicates a scenario where a tweet appears positive on the surface but is implicitly negative (hence, sarcastic). $z$ is a tweet-level latent variable, denoting the topic of the tweet. The number of topics, $Z$ is experimentally determined. $is$ is a word-level latent variable representing if a word is a topic word or a sentiment word, similar to  ~\newcite{mukherjee2012modeling}. If the word is a sentiment word, the word-level latent variable $s$ represents the sentiment of that word. It can take $S$ unique values. Intuitively, $S$ is set as $2$.

Among the distributions, $\eta_{w}$ is an observed distribution that is estimated beforehand. It denotes the probability of the word $w$ being a topic word or a sentiment word. Distribution $\theta_{l}$ represents the distribution over $z$ given the label of the tweet as $l$. $\psi_{l}$ and $\psi_{zl}$ are an hierarchical pair of distributions. $\psi_{zl}$ represents the distribution over sentiment of the word given the topic and label of the tweet and that the word is a sentiment word. $\chi_s$ and $\chi_{sz}$ are an hierarchical pair of distributions , where $\chi_{sz}$ represents distribution over words, given the word is a sentiment word with sentiment $s$ and topic $z$. $\phi_z$ is a distribution over words given the word is an topic word with topic $z$. The generative story of our model is:
{
\small
\textit{
\begin{enumerate}\setlength{\topsep}{0pt}\setlength{\itemsep}{0pt}
\setlength\itemsep{0cm}
 \item For each label $l$, select
 $~~~~~\vec\theta_l \!\sim\! \textrm{Dir}(\alpha)$ 
\item For each label $l$, select 
 $~~~~~\vec\psi_{l} \!\sim\! \textrm{Dir}(\beta_1)$\\
 $~~~~~$For each topic $z$, select\\
 $~~~~~~~~~~\vec\psi_{l,z}\!\sim\!\textrm{Dir}(\beta_2\vec\psi_{l})$
 \item For each topic $z$ and sentiment $s$, select 
 $~~~~~~~~~~\vec\chi_s\!\sim\! \textrm{Dir}(\delta_1)$, and 
 $\vec\chi_{s,z}\!\sim\! \textrm{Dir}(\delta_2\vec\chi_{s})$
\item For each topic $z$ select
  $~~~~~\vec\phi_z\!\sim\!\textrm{Dir}(\gamma)$
\item For each tweet $k$ select
\begin{enumerate}\setlength{\topsep}{0pt}\setlength{\itemsep}{0pt}
\vspace{-.3em}
 \item topic $z_k \sim \vec\theta_{l_k}$ 
 \item switch value for all words, $is_{k_j} \sim \vec\eta_j$
\item sentiment for all sentiment words, $s_{kj} \sim \vec\psi_{z_k,l_k}$
 \item all topic words, $w_{kj} \sim \vec\phi_{z_k}$
 \item all sentiment words, $w_{kj} \sim \vec\chi_{s_{kj},z_k}$
 \end{enumerate}
\end{enumerate}
}
}
We estimate these distribution using Gibbs sampling. The joint probability over all variables is decomposed into these distributions, based on dependencies in the model. Estimation details have not been included due to lack of space.

\section{Experiment Setup}
\label{sec:expsetup}
We create a dataset of English tweets for our topic model. We do not use datasets reported in past work (related to classifiers) because topic models typically require larger datasets than classifiers. The tweets are downloaded from twitter using the twitter API\footnote{https://dev.twitter.com/rest/public} using hashtag-based supervision. Hashtag-based supervision is common in sarcasm-labeled datasets~\cite{joshiacl}. Tweets containing hashtags \#happy, \#excited are labeled as positive tweets. Tweets with  \#sad, \#angry are labeled as negative tweets. Tweets with \#sarcasm and \#sarcastic are labeled as sarcastic tweets. The tweets are converted to lowercase, and the hashtags used for supervision are removed. Function words\footnote{www.sequencepublishing.com}, punctuation, hashtags, author names and hyperlinks are removed from the tweets. Duplicate tweets (same tweet text repeated for multiple tweets) and re-tweets (tweet text with the `RT' added in the beginning) are discarded. Finally, words which occur less than three times in the vocabulary are also removed. As a result, the tweets that have less than 3 words are removed. This results in a dataset of 166,955 tweets. Out of these, 70,934 are positive, 20,253 are negative and the remaining 75,769 are sarcastic. A total of 35398 tweets are used for testing, out of which 26,210 are of positive sentiment, 5535 are of negative sentiment and 3653 are sarcastic. We repeat that these labels are determined based on hashtags, as stated above.

The total number of  distinct labels ($L$) is 3, and the total number of distinct sentiment ($S$) is 2. The total number of distinct topics ($Z$) is experimentally determined as 50. We use block-based Gibbs sampling to estimate the distributions. The block-based sampler samples all latent variables together based on their joint distributions. We set asymmetric priors based on sentiment word-list from \newcite{mcauley2013amateurs}.

A key parameter of the model is $\eta_w$ since it drives the split of a word as a topic or a sentiment word. SentiWordNet \cite{baccianella2010sentiwordnet} is used to learn the distribution $\eta_w$ prior to estimating the model. We average across multiple senses of a word. Based on the SentiWordNet scores to all senses of a word, we determine this probability. 

\section{Results}
\label{sec:res}


\begin{table*}[h]
\centering
\small
\begin{tabular}{llll}
\toprule
\multicolumn{1}{c}{\textbf{Work}} & \multicolumn{1}{c}{\textbf{Party}} & \multicolumn{1}{c}{\textbf{Jokes}} & \multicolumn{1}{c}{\textbf{Weather}} \\ \midrule
day & life & Quote & Snow \\ 
morning & friends & Jokes & Today \\ 
night & night & Humor & Rain \\ 
today & drunk & Comedy & Weather \\ 
work & parties & Satire & Day \\ \midrule
\textbf{Women} & \multicolumn{1}{c}{\textbf{School}} & \multicolumn{1}{c}{\textbf{Love}} & \textbf{Politics} \\ \midrule
Women & tomorrow & love & Ukraine \\ 
Wife & school & feeling & Russia \\ 
Compliment(s) & work & break-up & again \\ 
Fashion & morning & day/night & deeply \\ 
Love & night & sleep & raiders \\ \midrule
\end{tabular}
\caption{Topics estimated when the topic model is learned on only sarcastic tweets}
\label{tab:sarctopics1}
\end{table*}
\subsection{Qualitative Evaluation}
The goal of this section is to present topics discovered by our sarcasm topic model. We do so in two steps. We first describe the topics generated when only sarcastic tweets from our corpus are used to estimate the distributions, followed by the ones when the full corpus is used. In case of the former, since only sarcastic tweets are used, the topics generated here indicate words corresponding to sarcasm-prevalent topics. In case of the latter, the sentiment-topic distributions in the model capture the prevalence of sarcasm.

The model estimates the $\phi$ and $\chi$ distributions corresponding to topic words and sentiment words. Top five words for a subset of topics (as estimated by $\phi$) are shown in Table~\ref{tab:sarctopics1}. The headings in boldface are manually assigned\footnote{This is a common practice in topics model papers, in order to interpret topics.~\cite{mukherjee2012aspect,joshi2016political,kim2013hierarchical}}. Sarcasm-prevalent topics, as discovered by our topic model, are work, party, weather, women, etc. The corresponding sentiment topics for each of these sarcasm-prevalent topics  (as estimated by $\chi$) are given in Table~\ref{tab:sarctopics2}. The headings in boldface are manually assigned. For topics corresponding to `party' and `women', we observe that the two columns contain words from opposing sentiment polarities. An example sarcastic tweet about work is `Yaay! Another night spent at office! I love working late night'. 

\begin{table*}[h]
\centering
\small
\begin{tabular}{llllllll}
\hline
\multicolumn{2}{c}{\textbf{Work}} & \multicolumn{2}{c}{\textbf{Party}} & \multicolumn{2}{c}{\textbf{Jokes}} & \multicolumn{2}{c}{\textbf{Weather}} \\ \hline
 Love & Great & Lol &  Hate & Funny & Lol & Love & nice \\ 
Good  & Sick &Attractive  &  Allergic & Liar & Fucks & Glad & wow \\ 
 Awesome & Seriously & Love  & Insulting & Hilarious & Like & Fun & really \\ \hline
\multicolumn{2}{c}{\textbf{Women}} & \multicolumn{2}{c}{\textbf{School}} & \multicolumn{2}{c}{\textbf{Love}} & \multicolumn{2}{c}{\textbf{Politics}} \\ \hline
Compliment(s) & Talents & excited & fun &  best & love & Losing & issues \\ 
Thrilled & Sorry & love & omg & awesome & ignored  & lies& weep \\ 
Recognized & Bad & really & awesome & greatest  & sick & like& really\\ \hline
\end{tabular}
\caption{Sentiment-related topics estimated when the topic model is learned on only sarcastic tweets}
\label{tab:sarctopics2}
\end{table*}

\begin{table*}[h]
\centering
\small
\begin{tabular}{llllll} \toprule
\textbf{Music} & \textbf{work/school} & \textbf{\begin{tabular}[c]{@{}l@{}}Orlando \\ Incident\end{tabular}} & \textbf{Holiday} & \textbf{Quotes} & \textbf{Food} \\ \midrule
pop & work & orlando & summer & quote(s) & food \\
country & sleep & shooting & wekend & morning & lunch \\
rock & night & prayers & holiday & inspiration & vegan \\
bluegrass & morning & families & friends & motivation & breakfast \\
beatles & school & victims & sun,beach & mind & cake \\ \midrule
\end{tabular}
\begin{tabular}{llllll}
\textbf{\begin{tabular}[c]{@{}l@{}}Stock(s)/\\ Commodities\end{tabular}} & \textbf{Father} & \textbf{Gun}  & \textbf{Pets} & \textbf{Health} \\ \midrule
silver & father(s) & gun(s)  & dog & fitness \\
gold & dad & orlando  & cat & gym \\
index & daddy & trump  & baby & run \\
price & family & shooting  & puppy & morning \\
consumer & work & muslim  & pets & health \\ \bottomrule
\end{tabular}
\caption{Topics estimated when the topic model is learned on full corpus}
\label{tab:fulltopics1}
\end{table*}
\begin{figure}
\centering
\includegraphics[width=0.50\textwidth]{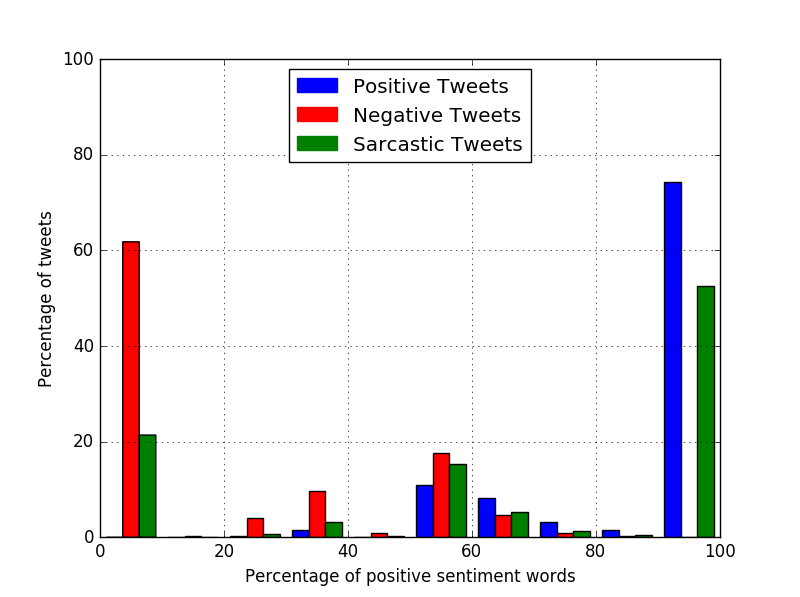}
\caption{Distribution of word-level sentiment labels for tweet-level labels}
\label{fig:graph}
\end{figure}
\begin{table*}[h]
\centering
\small
\begin{tabular}{p{1.25cm}lllllp{1cm}p{1cm}llll}\toprule
\multicolumn{2}{c}{\textbf{\begin{tabular}[c]{@{}c@{}}Stock(s)/\\ Commodities\end{tabular}}} & \multicolumn{2}{c}{\textbf{Father}} & \multicolumn{2}{c}{\textbf{Gun}}  & \multicolumn{2}{c}{\textbf{Pets}} & \multicolumn{2}{c}{\textbf{Health}} & \multicolumn{2}{c}{\textbf{Music}} \\ \midrule
gains & risks & happy & lol & like & sad   & happy & small & love & tired & love & sad\\
happiness & dipped & love & little & good & hate   & love & sad & fun & sick & happy & passion  \\
unchanged & down & best & bless & wow & angry   & cute & miss & laugh & unfit & good & pain \\ 
\end{tabular}
\begin{tabular}{p{1.25cm}lllllp{1cm}p{1cm}llll}\midrule
 \multicolumn{2}{c}{\textbf{Work/School}} & \multicolumn{2}{c}{\textbf{\begin{tabular}[c]{@{}c@{}}Orlando\\ Incident\end{tabular}}} & \multicolumn{2}{c}{\textbf{Holiday}} & \multicolumn{2}{c}{\textbf{Quotes}} & \multicolumn{2}{c}{\textbf{Food}} \\ \midrule
 great & sick & love & tragedy & love & beauty & positive & simple & happy & foodie \\
 fun & hate & like & hate & smile & hot & happy & kind & yummy & seriously \\ 
 yay & ugh & want & heartbroken & fun & sexy & happiness & sad & healthy & perfect\\ \bottomrule
\end{tabular}
\caption{Sentiment-related topics estimated when the topic model is learned on full corpus}
\label{tab:fulltopics2}
\end{table*}
The previous set of topics are all from sarcastic text. We now show the topics extracted by our model from the full corpus. These topics will indicate peculiarity of topics for each of the three labels, allowing us to infer what topics are sarcasm-prevalent. Table~\ref{tab:fulltopics1} shows the top 5 topic words for the topics discovered (as estimated in $\phi$) from the full corpus (\textit{i.e.}, containing tweets of all three tweet-level sentiment labels: positive, negative and sarcastic). Table~\ref{tab:fulltopics2} shows the top 3 sentiment words for each sentiment  (as estimated by $\chi$) of each of the topics discovered. Like in the previous case, the heading in boldface is manually assigned. One of the topic discovered was `Music'. The top 5 topic words for the topic `Music' are Pop, Country, Rock, Bluegrass and Beatles.  The corresponding sentiment words for Music are `love', `happy', `good' on the positive side and `sad', `passion' and `pain' on the negative side. 

The \textbf{remaining sections present results when the model is learned on the full corpus}.



\subsection{Quantitative Evaluation}
In this section, we answer three questions: (A) What is the likely sentiment label, if a user is talking about a particular topic? (Section 6.2.1), (B) We hypothesize that sarcastic text tends to have mixed-polarity words. Does it hold in case of our model? (Section 6.2.2), and (C) How can sarcasm topic model be used for sarcasm detection? (Section 6.2.3).

\subsubsection{Probability of sentiment label, given topic}
We compute the probability $p(l/z)$ based on the model. 
\begin{table}[]
\centering
\small
\begin{tabular}{llll}
\toprule
\textbf{Topics P(l/z)} & \textbf{Positive} & \textbf{Negative} & \textbf{Sarcastic} \\ \midrule
Holiday & \textbf{0.9538} & 0.0140 & 0.0317 \\
Father & \textbf{0.9224} & 0.0188 & 0.0584 \\
Quote & \textbf{0.8782} & 0.0363 & 0.0852 \\
Food & \textbf{0.8100} & 0.0331 & 0.1566 \\
Music & \textbf{0.7895} & 0.0743 & 0.1363 \\
Fitness & \textbf{0.7622} & 0.0431 & 0.1948 \\
Orlando  Incident & 0.0130 & \textbf{0.9500} & 0.0379 \\
Gun & 0.1688 & 0.3074 & \textbf{0.5230} \\
Work & 0.1089 & 0.0354 & \textbf{0.8554}\\
Humor & 0.0753 & 0.1397 & \textbf{0.7841} \\ \bottomrule
\end{tabular}
\caption{Probability of sentiment label for various discovered topics}
\label{tab:labelgiventopics}
\end{table}
Table~\ref{tab:labelgiventopics} shows these values for a subset of topics. Topics with a majority positive sentiment are Father's Day (0.9224), holidays (0.9538), etc. The topic with the highest probability of a negative sentiment is the Orlando shooting incident (0.95). Gun laws (0.5230), work and humor are where sarcasm is prevalent.
\subsubsection{Distribution of sentiment words for tweet-level sentiment labels}
Figure~\ref{fig:graph} shows the proportion of word-level sentiment labels, for the three tweet-level sentiment labels, \textbf{as estimated by our model}. The X-axis indicates percentage of positive sentiment words in a tweet, while Y-axis indicates percentage of tweets which indicate a specific value of percentage. More than 60\% negative tweets (bar in red) have 0\% positive content words. The `positive' here indicates the value of $s$ for a word in a tweet. In other words, the said red bar indicates that 60\% tweets have 0\% words sampled with $s$ as positive.

It follows intuition that negative tweets have low percentage of positive words (red bar on the left part of the graph) while positive tweets have high percentage of positive words (blue bar on the right part of the graph). The interesting variations are observed in case of sarcastic tweets. It must be highlighted that ~\textit{the sentiment labels considered for these proportions are \textbf{as estimated by our topic model}}. Many sarcastic tweets contain very high percentage of positive sentiment words. Similarly, the proportion of tweets with around 50\% positive sentiment words is around 20\%, as expected. Thus, the model is able to capture the sentiment mixture as expected in the three tweet-level sentiment labels: (literal) positive, (literal) negative and sarcastic.

\subsubsection{Application to Sarcasm Detection}
We now use our sarcasm topic model to detect sarcasm, and compare it with two prior work. The task here is to classify a tweets as either sarcastic or not. We use the topic model for sarcasm detection using two methods:
\begin{enumerate}\setlength{\topsep}{0pt}\setlength{\itemsep}{0pt}
    \item \textbf{Log-likelihood based}: The topic model is first learned using the training corpus where the distributions in the model are estimated. Then, the topic model performs sampling for a pre-determined number of samples, in three runs - once for each label. For each run, the log-likelihood of the tweet given the estimated distributions (in the training phase) and the sampled values of the latent variables (for this tweet) is computed. The label of  the tweet is returned as the one with the highest log-likelihood.
    \item \textbf{Sampling-based}: Like in the previous case, the topic model first estimates distributions using the training corpus. Then, the topic model is learned again where the label $l$ is assumed to be latent, in addition to the tweet-level latent variable $z$, and word-level latent variables $s$, and $is$. The value of $l$ as learned by the sampler is returned as the predicted label.
\end{enumerate} 
We compare our results with two previously existing techniques, \newcite{buschmeier2014impact} and \newcite{liebrecht2013perfect}. We ensure that our implementations result in performance comparable to the reported papers. The two rely on designing sarcasm-level features, and training classifiers for these features. For these classifiers, the positive and negative labels are combined as non-sarcastic. As stated above, the test set is separate from the training set. The results of these two past methods compared with the two based on topic models are shown in Table~\ref{tab:sarcdetect}. Both prior work show poor F-score (around 18-19\%) while our log-likelihood based approach achieves the best F-score of 41.34\%. The low values, in general, may be because our corpus is large in size, and is diverse in terms of the topics. Also, features in ~\newcite{liebrecht2013perfect} are unigrams, bigrams and trigrams which may result in sparse features.

\begin{table}[h]
\centering
\small
\begin{tabular}{llll}
\toprule
\textbf{Approach} & \textbf{P (\%)} & \textbf{R (\%)} & \textbf{F (\%)} \\ \midrule
\cite{buschmeier2014impact} & 10.41 & 100.00 & 18.85 \\
\cite{liebrecht2013perfect}& 11.03 & 99.88 & 19.86 \\
Topic Model: Log Likelihood& 55.70 & 32.87 & \textbf{41.34} \\
Topic Model: Sampling& 58.58 & 13.11 & 21.42\\ \bottomrule
\end{tabular}
\caption{Comparison of Various Approaches for Sarcasm Detection}
\label{tab:sarcdetect}
\end{table}


\section{Conclusion \& Future Work}
\label{sec:concl}
We presented a novel topic model that discovers sarcasm-prevalent topics. Our topic model uses a dataset of tweets (labeled as positive, negative and sarcastic), and estimates distributions corresponding to prevalence of a topic, prevalence of a sentiment-bearing words. We observed that topics such as work, weather, politics, etc. were discovered as sarcasm-prevalent topics. We evaluated the model in three steps: (a) Based on the distributions learned by our model, we show the most likely label, for all topics. This is to understand sarcasm-prevalence of topics when the model is learned on the full corpus. (b) We then show distribution of word-level sentiment for each tweet-level sentiment label as estimated by our model. \textit{Our intuition that sentiment distribution in a tweet is different for the three labels: positive, negative and sarcastic, holds true}. (c) Finally, we show how topics from this topic model can be harnessed for sarcasm detection. We implement two approaches: one based on most likely label as per log likelihood, and another based on last sampled value during iteration. In our log-likelihood based approach, we are able to significantly outperform two prior work based on feature design by F-Score of around 25\%.

The current model is limited because of its key intuition about sentiment mixture in sarcastic text. Instances such as hyperbolic sarcasm go against the intuition. The current approach relies only on bag of words which may be extended to n-grams since a lot of sarcasm is expressed through phrases with implied sentiment. This work, being an initial work in topic models for sarcasm, sets up the promise of topic models for sarcasm detection, as also demonstrated in corresponding work in aspect-based sentiment analysis.
\bibliographystyle{eacl2017}
\bibliography{references}
\end{document}